\pdfoutput=1

\documentclass[11pt]{article}

\usepackage[]{EMNLP2023}

\usepackage{times}
\usepackage{latexsym}

\usepackage[T1]{fontenc}

\usepackage[utf8]{inputenc}

\usepackage{microtype}

\usepackage{inconsolata}

\usepackage{url}
\usepackage{hyperref}
\usepackage{algorithm}
\usepackage{algorithmic}
\usepackage{listings}
\usepackage{newfloat}
\usepackage{listings}
\usepackage{titlesec}
\usepackage{verbatim}
\usepackage{multirow}
\usepackage{graphicx}
\usepackage{booktabs}
\usepackage{amssymb}
\usepackage{amsmath}
\usepackage{bm}
\usepackage{color, soul}
\usepackage{color, soul, colortbl}

\usepackage[capitalise]{cleveref}

\usepackage[algo2e]{algorithm2e} 
\usepackage[textsize=tiny]{todonotes}

\definecolor{wingreen}{rgb}{0,0.45,0.24}
\definecolor{losered}{rgb}{1.0,0.1,0.24}

\newcommand{\methodname}[1]{DetectGPT4Code}
\newcommand{\basea}[1]{base}
\newcommand{\baseb}[1]{base}
\newcommand{\data}[1]{data}

\definecolor{bubbles}{rgb}{0.91, 1.0, 1.0}
\definecolor{losered}{rgb}{1.0,0.1,0.24}
\definecolor{lightgray2}{rgb}{0.8,0.8,0.8}
\newcommand{\doneWhite}{\cellcolor{bubbles}}
\newcommand{\done}{\cellcolor{lightgray2}}

\title{ Zero-Shot Detection of Machine-Generated Codes}

\author{Xianjun Yang$^{1}$ \qquad Kexun Zhang$^{1}$ \qquad Haifeng Chen$^{2}$ \\ \bf
Linda Petzold$^{1}$ \qquad William Yang Wang$^{1}$ \qquad Wei Cheng$^{2}$ \\
\texttt{\{xianjunyang, kexun, petzold, wangwilliamyang\}@ucsb.edu}\\ \texttt{\{weicheng, haifeng\}@nec-labs.com}\\
$^{1}$ University of California, Santa Barbara \qquad $^{2}$ NEC Laboratories America, Princeton}

\begin{document}
\maketitle

\begin{abstract}

This work proposes a training-free approach for the detection of LLMs-generated codes, mitigating the risks associated with their indiscriminate usage. 
To the best of our knowledge, our research is the first to investigate zero-shot detection techniques applied to code generated by advanced black-box LLMs like ChatGPT.
Firstly, we find that existing training-based or zero-shot text detectors are ineffective in detecting code, likely due to the unique statistical properties found in code structures.
We then modify the previous zero-shot text detection method, DetectGPT \citep{mitchell2023detectgpt} by utilizing a surrogate white-box model to estimate the probability of the rightmost tokens, allowing us to identify code snippets generated by language models. Through extensive experiments conducted on the python codes of the CodeContest and APPS dataset, our approach demonstrates its effectiveness by achieving state-of-the-art detection results on \texttt{text-davinci-003}, \texttt{GPT-3.5}, and \texttt{GPT-4} models.
Moreover, our method exhibits robustness against revision attacks and generalizes well to Java codes. We also find that the smaller code language model like \texttt{PolyCoder-160M} performs as a universal code detector, outperforming the billion-scale counterpart. The codes will be available at \url{https://github.com/Xianjun-Yang/Code\_detection.git}

\end{abstract}

\section{Introduction}

The scale-up of model and data size has led to influential AI products such as ChatGPT \citep{schulman2022chatgpt}, which excels in generating both text and code. Meanwhile, code LLMs like Codex \citep{chen2021evaluating} and AlphaCode \citep{li2022competition} have the capability to translate docstrings to Python functions and even solve competition-level programming problems. As the world's pioneering AI developer tool at scale, GitHub Copilot has revolutionized developer productivity for over one million individuals, enabling them to code up to 55\% faster \footnote{https://github.blog/2023-02-14-github-copilot-for-business-is-now-available/}. There is also increasing focus on developing code LLMs \citep{ nijkamp2022codegen, allal2023santacoder} for various tasks like multilingual code generation \citep{zheng2023codegeex}, code infilling \citep{fried2022incoder}. Furthermore, versatile text generation models like GPT-4 \citep{openai2023gpt4} possess the capability to solve programming tasks and have even sparked speculation about the potential development of artificial general intelligence \citep{bubeck2023sparks}. 

However, the risks also arise along with the widely used AI tools. Previous concerns about the risk of AI-generated text include misinformation spread \citep{bian2023drop, hanley2023machine, pan2023risk}, fake news \citep{oshikawa2018survey, zellers2019defending, dugan2022real}, gender bias \citep{sun-etal-2019-mitigating}, education \citep{perkins2023game}, social harm \citep{kumar-etal-2023-language}. Nonetheless, scant attention has been directed towards the realm of machine-generated code detection, perhaps owing to the subpar quality exhibited by previous small models. However, over the past two years, there has been a noteworthy advancement in the code-generation capabilities of LLMs, as highlighted in a survey \citep{zan2022neural}, which has consequently made the reliable detection of machine-generated codes a pressing concern. The potential harms associated with code generation encompass jeopardizing programming education, generating malicious software, and encroaching upon intellectual property protection. Given the broader impact of code generation, we firmly assert that the field of detection remains significantly underexplored and merits prompt and thorough investigation.

Recently, there has been an increasing focus on detecting ChatGPT-generated text using watermarks \citep{kirchenbauer2023watermark}, zero-shot \citep{ mitchell2023detectgpt, yang2023dna} or training-based  approaches \citep{AITextClassifier}, but it is still unclear whether those popular detectors can perform well on code contents. As demonstrated by \citep{wang2023evaluating}, previous detectors have exhibited lower performance when applied to code-related tasks such as code summarization and code generation.
It is also demonstrated that applying text watermark \citep{kirchenbauer2023watermark} to codes degrades both the code quality and detectability. The reason behind this is that codes differ in nature from text, typically with lower entropy. Drawing upon these observations, our objective is to evaluate text detection algorithms on code and introduce novel methodologies for detecting codes generated by state-of-the-art black-box LLMs like \texttt{GPT-4}.
 
\begin{figure*}
\centering
    \includegraphics[width=0.98\textwidth]{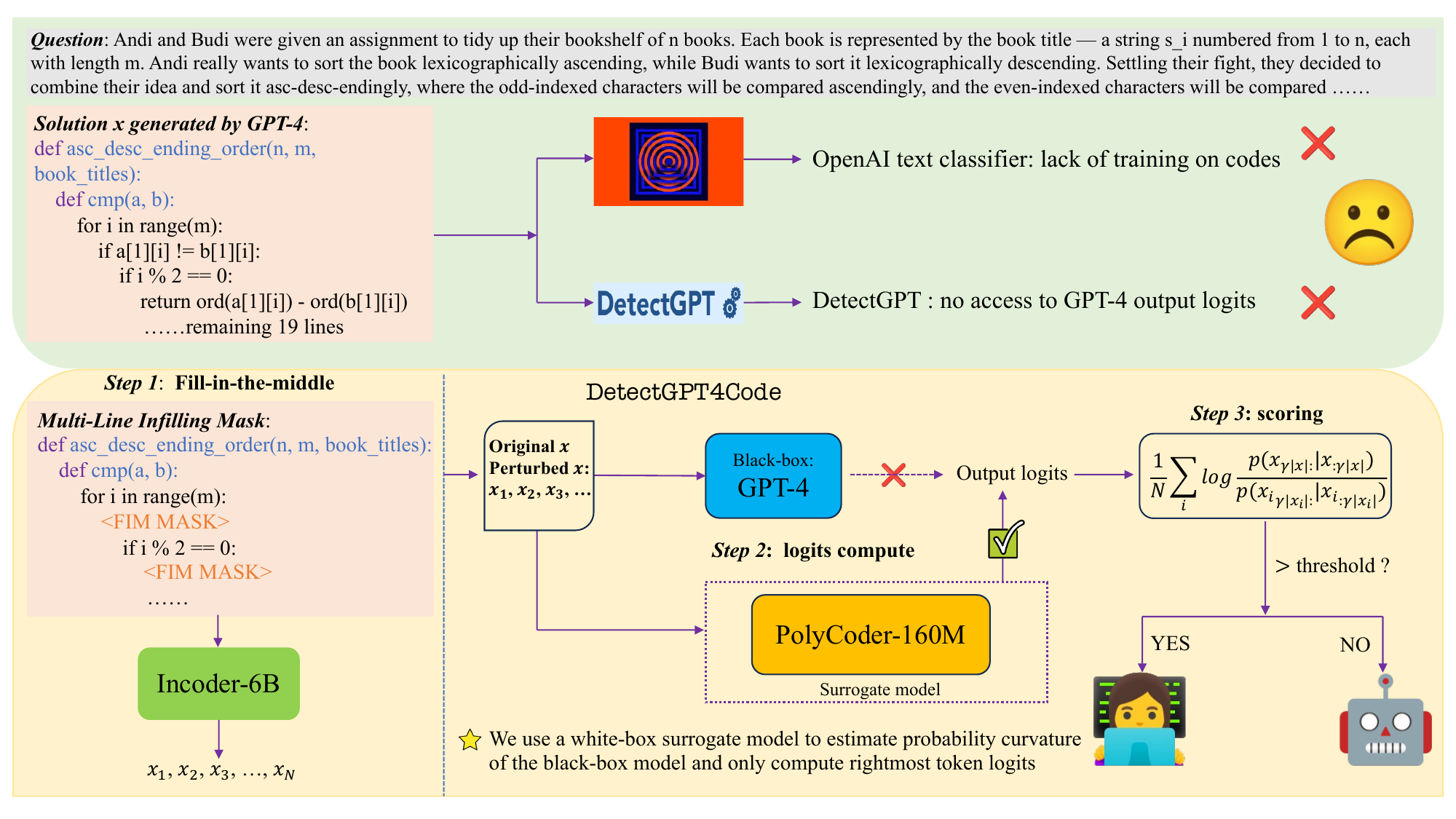}
    \caption{An overview of our framework. In the original DetectGPT, the output logits are needed for computation. But GPT-4 does not provide such logits, so we use a surrogate model to replace it for scoring by getting perturbed codes through filling-in-the-middle tasks. Then, we only calculate the rightmost token probabilities given preceding codes, controlled by a ratio $\gamma$. Finally, the decision is made by comparing the score with the pre-defined threshold. }\label{fig: overall}
\end{figure*}

In this study, our primary focus lies in achieving \textit{zero-shot detection of codes generated from black-box LLMs}. Firstly, we evaluate existing detection tools across different LLMs and datasets to demonstrate their shortcomings. Subsequently, we propose novel detection approaches and conduct extensive experiments to assess their effectiveness. The overall framework of our method is shown in Figure \ref{fig: overall}. 
To summarize, our work provides three main contributions:
\begin{itemize}
    \item
    We are the first to systematically demonstrate the failure of previous popular text detectors in code detection, encompassing a diverse range of LLMs and benchmarks.
    \item 
    We propose a novel detection method specifically designed for code generated by black-box LLMs, surpassing previous baselines by a considerable margin. Furthermore, we validate its robustness against revision attacks, highlighting its effectiveness in detecting revised code.
    \item 
    Our findings reveal that smaller code language models serve as superior universal code detectors, with \texttt{PolyCoder-160M} emerging as the most effective surrogate detector across various models, datasets, and programming languages.
    
\end{itemize}

\section{Related Work}

\subsection{LLMs for Code Generation}
Code LLMs, exemplified by pioneering works such as CodeX \citep{chen2021evaluating}, AlphaCode \citep{li2022competition}, and general models like GPT-4 \citep{openai2023gpt4}, have been purposefully crafted for code generation tasks, showcasing remarkable capabilities in generating high-quality code. Besides, CodeGeex \citep{zheng2023codegeex} is a 13 billion model trained for code generation with multilingual evaluations on their newly curated humaneval-x benchmark.
SantaCoder \citep{allal2023santacoder} surpasses previous open-source multilingual code generation models and advises against removing code repositories with low stars on GitHub. In the meanwhile, InCoder6.7B \citep{fried2022incoder} and CodeGen-Multi-2.7B \citep{nijkamp2022codegen}) are both trained by left-to-right generation and can perform code infilling on the Java, JavaScript, and Python. Recently, WizardCoder \citep{luo2023wizardcoder} significantly outperforms other open-source Code LLMs by employing the Evol-Instruct method for complex instruction fine-tuning.
Subsequently, numerous endeavors have been undertaken to enhance the quality of code generation \citep{ni2023lever, zhang2023algo, chen2023teaching} by new decoding strategies. 

\subsection{Detection of Machine-Written Codes}
While there has been a growing emphasis on detecting text generated by LLMs, to the best of our knowledge, there has been limited exploration into the detection of LLMs-generated code. \citep{lee2023wrote} has shown that conventional text detectors like DetectGPT \citep{mitchell2023detectgpt} struggle to effectively detect codes. Additionally, they found that the perturbation model T5 \citep{raffel2020exploring}, originally designed for code perturbation, underperformed and was replaced with the expert code generation model, SantaCoder \citep{allal2023santacoder}. Furthermore, it has been observed by \citep{kirchenbauer2023watermark} that conventional watermarking techniques designed for text also prove ineffective when applied to code, and that code quality, as measured by pass@k \citep{chen2021evaluating}, deteriorates. To address this, they propose a straightforward modification by excluding low entropy tokens from watermarking, which leads to significant improvements. However, it is important to note that this watermarking approach still necessitates access to the full model weights, rendering it ineffective for black-box models. 
Another notable area of research revolves around software watermarking, which primarily aims to protect the intellectual property of commercial software programs. This field encompasses static software watermarking methods that involve embedding watermarks into code using techniques such as code replacement and re-ordering \citep{hamilton2011survey, myles2005evaluation}. Additionally, dynamic software watermarking techniques involve injecting watermarks during the compilation/execution stage \citep{wang2018exception, dey2019software, ma2019xmark}. However, those watermarking are not designed to distinguish between humans and LLMs.

\section{Detection Formulation}

We conduct extensive experiments on three models from OpenAI and 2 datasets covering Python and Java to evaluate the code detection performance. In this section, we briefly discuss more details about the used models, datasets, baselines and our novel methods.
\subsection{Datasets and Models}

\textbf{Datasets.}
We consider two challenging code generation datasets. 
CodeContests \citep{li2022competition} is a competitive programming dataset used for training AlphaCode \citep{li2022competition}. We use the test split of 165 samples on Python and Java. The human-written answers are on average 1652 characters. We discard long answers with more than 3500 characters as they exceed the context window of some models. This left us with 102 test instances. We denote the resulting corpus as CodeContest-Python and CodeContest-Java.
Another dataset is the Automated Programming Progress Standard, abbreviated APPS \citep{hendrycks2021measuring}, curated from open-access sites and filtered for quality control. It includes Introductory, Interview and Competition Level difficulties. We choose the 135 competition level problems in the test set. The average length of characters and lines is 470 and 21, respectively. 

\textbf{Code LLMs.} We test the detection on the most advanced LLMs from OpenAI: \texttt{GPT-4}, \texttt{GPT-3.5-turbo}, and \texttt{text-davinci-003}, as they already have been widely used by more than 100 million users, especially students. We mainly focus on black-box detection to simulate the realistic setting. We use the API service \footnote{https://openai.com/blog/openai-api} from OpenAI. The decoding temperature is set to $0.7$, with a maximum of $1024$ new tokens during generation, while all other parameters remain at their default values. 
All experiments are conducted within the timeframe of May 23 to June 23, 2023, and all solutions are obtained by prompting the original questions into the LLMs. To account for instances where the returned answers contain additional text beyond the code, we adopt the approach outlined in \citep{chen2021evaluating} to filter out such extraneous text. The remaining filtered codes are subsequently considered as machine-generated codes for our analysis.

\subsection{Metrics} 
We mainly use the Area Under the Receiver Operating Characteristic (AUROC) score to evaluate the effectiveness of detection algorithms. Furthermore, in accordance with the recommendations put forth by \citep{krishna2023paraphrasing, yang2023dna}, we prioritize maintaining a high True Positive Rate (TPR) while minimizing the False Positive Rate (FPR). Specifically, we report TPR scores at a fixed 10\% FPR. We opted not to select a 1\% FPR threshold as the total number of instances is typically limited (around 100), and allowing only one error would be excessively stringent.

\subsection{Baselines}
Due to the relatively unexplored nature of AI-generated code detection, the availability of code detectors is limited. In this study, we employ four baseline text classifiers, as outlined below. More details can be found in Appendix \ref{app: baseline}.

\textbf{Training-based Baselines.} We consider OpenAI's text classifier \citep{AITextClassifier} and GPTZero \citep{GPTZero} as training-based baselines since they all claim that their classifiers are trained on millions of text and can generalize to many different models from various organizations. Following \citep{krishna2023paraphrasing}, we use the \texttt{model-detect-v2} as OpenAI's text classifier and GPTZero \footnote{https://gptzero.me/docs} both through their API services.

\textbf{Zero-shot Baselines.} We also consider the original DetectGPT \citep{mitchell2023detectgpt} and DNA-GPT \citep{yang2023dna} since they are both strong zero-shot detectors. However, only the BScore in DNA-GPT can be used for black-box detection, while both the canonical DetectGPT and WScore in DNA-GPT require access to the model output logits. Thus, we only use the white-box detection of DNA-GPT on \texttt{text-davinci-003} since it is the only available model providing partial output logits.
Notice that in the original DetectGPT paper, the authors investigate using a surrogate model to estimate the source model probability but show a large gap compared to using the source model itself.  \citep{mireshghallah2023smaller} claims OPT-125M is the best surrogate model and universal detector for cross-model detection using DetectGPT. However, it is important to highlight that these methods were only evaluated for text detection, raising doubts about their effectiveness when applied to code detection. We thus use their recommended settings for code detection. As suggested by \citep{lee2023wrote}, we replace the perturbation model in DetectGPT from T5-large \citep{raffel2020exploring} to code generation model, Incoder-6B \citep{fried2022incoder}, since T5 is not good at perturbed codes generation.

\section{Our Method}

In this section, we present our enhanced methodology for detecting black-box models, employing a white-box code LM as a surrogate model to estimate the probability curve. Building upon the foundations of DetectGPT \citep{mitchell2023detectgpt}, we have tailored our approach specifically for code detection, thus dubbing it DetectGPT4Code.

\subsection{DetectGPT4Code}
The original DetectGPT can not be directly applied to black-box detection since we do not have access to the model behind the wall. Instead, we propose to use a smaller code LM as a surrogate model for calculating the probability curve. Also, \citep{bavarian2022efficient} demonstrates efficient training of language models to fill-in-the-middle (FIM) under the single-line, multi-line, and random span infilling of codes. They found autoregressive model trained with the FIM objective can perform code infilling without harming the original left-to-right generative capability.  \\
\textbf{Modification 1: Perturbations Simulation.} In practice, OpenAI provides a text editing model like \texttt{text-davinci-edit-001} which shows good code editing ability, nevertheless it has a monetary cost, especially for a large number of Perturbations. Instead, we opt for a free open-sourced model for simulating perturbations. In particular, we use Incoder-6B \citep{fried2022incoder} due to its strong coding ability for the FIM task by first masking a span of $m$ lines of codes and then refilling them. Given the candidate code $c$, we denote the refilled codes as $\{c_1, ..., c_N\}$, where $N$ is the number of perturbations.

\begin{table*}[ht]
\centering
\resizebox{1.\textwidth}{!}
{
\begin{tabular}{lcccccccc}
\toprule
Datasets & \multicolumn{2}{c}{CodeContest-python} & \multicolumn{2}{c}{CodeContest-Java} & \multicolumn{2}{c}{APPS-python} & \multicolumn{2}{c}{Average} \\
\cmidrule(lr){2-3} \cmidrule(lr){4-5} \cmidrule(lr){6-7} \cmidrule(lr){8-9}  
Method & AUROC & TPR & AUROC & TPR & AUROC & TPR & AUROC & TPR  \\
\midrule
\multicolumn{9}{c}{\done  \texttt{GPT-4-0314} }\\
\midrule
GPTZero \cite{GPTZero} & 67.68 & 4.10 & 43.72 & 3.54 & 65.99 & 26.67 & 59.13 & 11.44 \\
OpenAI \cite{AITextClassifier} & 48.34  & 12.30 & \textbf{84.16} & 36.28 & 73.95 & 14.07 & 68.78 & 20.88 \\
\hline
DetectGPT \cite{mitchell2023detectgpt} & 54.01 & 19.81 & 54.89 & 18.75 & 61.78 & 28.89 & 56.89 & 22.48 \\
DNA-GPT \cite{yang2023dna} & 79.16 & 29.75 & 64.25 & 5.31 & \textbf{86.67} & \textbf{57.03} & 76.70 & 30.70 \\
\hline
Our DetectGPT4Code, $N$=20, $\gamma$=0.9 & \textbf{86.01} & \textbf{62.75} & \textbf{76.25} & \textbf{52.59} & 76.25 & 52.29 & \textbf{79.50} & \textbf{55.88} \\

\midrule
\multicolumn{9}{c}{\done  \texttt{GPT-3.5-turbo}} \\
\midrule

GPTZero & 58.69 & 21.08 & 31.65 & 7.08 & 67.85 & 35.56 & 52.73 & 21.24 \\
OpenAI & 64.21 & 17.65 & 42.06 & 2.65 & 46.58 & 4.44 & 50.95 & 8.25 \\
\hline
DetectGPT & 48.73 & 1.90 & 55.12 & 9.73 & 53.02 & 14.81 & 52.29 & 8.81 \\
DNA-GPT & 71.08 & \textbf{24.37} & 61.22 & 10.62 & 73.39 & 28.15 &  68.56 & 21.05 \\
\midrule
Our DetectGPT4Code, $N$=20, $\gamma$=0.99 & \textbf{71.46} & 20.01 & \textbf{64.03} & \textbf{28.32} & \textbf{78.05} & \textbf{44.44} & \textbf{71.18} & \textbf{30.92} \\

\midrule
\multicolumn{9}{c}{\done \texttt{text-davinci-003} } \\
\midrule
GPTZero  & 71.41 & 38.52 & 42.89 & 5.36 & 69.02 & 29.55 & 61.11 & 24.48 \\
OpenAI   & 23.81  & 4.10 & 27.78 & 4.46 & 27.19 & 3.03 & 26.26 & 3.86 \\
\hline
DetectGPT   & 46.81 & 11.32 & 54.30 & 8.93 & 58.93 & 20.66 & 53.35 & 13.64 \\
\done DNA-GPT-BScore & 77.49 & 4.13 & 71.60 & 4.46 & \textbf{80.19} & 14.39 & 76.43 & 7.66 \\
\doneWhite DNA-GPT-WScore & \textbf{78.89} & 8.09 & 69.79 & 2.70 & 71.24 & 37.88 & 73.31 & 16.22 \\
\midrule
Our DetectGPT4Code, $N$=20$, \gamma$=0.99 & 78.63 & \textbf{40.95} & \textbf{77.42} & \textbf{45.54} & 79.89 & \textbf{44.63} & \textbf{78.65} & \textbf{43.71} \\ 

\bottomrule
\end{tabular}

}
\caption{Overall comparison of different methods on three datasets. The TPR is calculated at 10\% FPR 
. $N$ in DetectGPT4Code represents the number of perturbations. BScore and WScore represent the black-box and white-box detection in their original DNA-GPT paper. All results are reported by percentage.
}\label{tab:overall} 
\end{table*}

\textbf{Modification 2: Surrogate Probability Calculation.} Since we are aiming at detecting black-box machine-generated codes, there is no direct model outputs probability for a closed model like \texttt{GPT-4}. Instead, we opt for another white-box model as a surrogate for accessing the probability curve.
We consider a series of smaller models including \texttt{PyCodeGPT-110M} \citep{CERT}, \texttt{PolyCoder-160M, 0.4B, 2.7B} \citep{xu2022systematic}, \texttt{CodeParrot-1.5B} \citep{codeparrot}, \texttt{LLaMa-13B} \citep{touvron2023llama}, ranging from around ~100M to 10B in parameters.
We do not consider other models with Encoder-Decoder architecture like CodeT5 \citep{wang-etal-2021-codet5}, PyMT5 \citep{clement-etal-2020-pymt5}, ERNIE-Code \citep{chai2022ernie} considering they might not be suitable for probability estimate of decoder-only models. 
The surrogate model is denoted as $LM_{surrogate}$, we then get the estimated probabilities: $\hat{p}_{original} = LM_{surrogate}(c)$, and $\{\hat{p}_{c_1} = LM_{surrogate}(c_1), ... , \hat{p}_{c_N} = LM_{surrogate}(c_N)\}$. 

\textbf{Modification 3: Less is More.}
Our preliminary experiments find that using the total probability of the full-length codes results in suboptimal performance. Therefore, we consider using fewer tokens as anchors to compare their probability divergence. The motivation is that: \textit{the beginning tokens usually have a lower probability, and thus their higher log probabilities will dominate the calculation. Instead, the ending tokens are more deterministic given enough preceding text and thus are better indicators.} The quantitative evidence to support this claim is shown in Figure \ref{fig: ratio} as will be discussed in the next section.
Correspondingly, we define $\gamma$ as the preceding text ratio and only compute the continuation code $c_{\gamma |c|: }$ probability given the preceding code $c_{:\gamma |c|}$. The resulting DetectGPT4Code score is calculated as follows:
$$ \hat{p}_{original}(c_{\gamma |c|: } | c_{:\gamma |c|}) - \frac{1}{N} \sum_{i=1}^N \hat{p}_{c_i}({c_i}_{\gamma |c_i|: } | {c_i}_{:\gamma |c_i|}).$$

Compared with the originally recommended words mask ratio $15\%$ in DetectGPT \citep{mitchell2023detectgpt}, we use the number of masking lines $m \in \{2, 4, 8, 12\}$ in our implementation. We perform perturbations on each instance for $N \in \{2, 4, 8, 20, 32, 64, 80, 100\}$ times to explore how many perturbations are required to achieve a good performance.

\subsection{Detector Models} For our detection models, we consider a range of 11 widely used open-access surrogate language models, varying in size from millions to billions of parameters: CodeParrot-110M, 1.5B \citep{codeparrot}, OPT-125M \citep{zhang2022opt}, GPT-neo-125M, 1.3B, 2.7B \citep{gao2020pile}, PolyCoder-125M, 0.4B, 2.7B \citep{xu2022systematic}, Incoder-1B, 6B \citep{fried2022incoder}.
We obtain all of them from Huggingface \footnote{https://huggingface.co/}. Except for OPT-125M, all other models are trained on a large collection of codes.

\section{Results and Analysis}

\subsection{Overall Results}
The overall results are shown in Table \ref{tab:overall}. As we can see, both training-based text detectors GPTZero and OpenAI's text classifier perform poorly on almost all datasets, sometimes with even worse results than a random guess. This indicates the vulnerability of those classifiers for code detection although codes are also LLMs-generated content, possibly due to a lack of training data. Besides, the previous SOTA zero-shot text classifier DetectGPT and DNA-GPT also demonstrate suboptimal performance when adapting to codes. The suspected reason behind this observation is that codes typically exhibit lower token entropy compared to text as evidenced in \citep{lee2023wrote} tested on LLaMa \citep{touvron2023llama} generated codes. However, we do not have access to the complete OpenAI model outputs for such validation.

On the contrary, our proposed code detection significantly improves the overall detection quality in terms of both AUROC and TPR scores across almost all models, programming languages, and datasets. We want to clarify that although the results are not as satisfactory as text detection, which usually achieves around 80 to 90 points of AUROC, the code detection is challenging and our method almost always outperforms all four baselines by a substantial margin. On average, our results demonstrate superior performance compared to previous methods, with improvements ranging from approximately 3 to 5 points in AUROC and 10 to 25 points in TPR.
Thus, we believe our approach can serve as a competitive competitor for future work.

\subsection{Influence of $\gamma$}
\begin{figure}[ht]
\centering
\includegraphics[width=1.\linewidth]{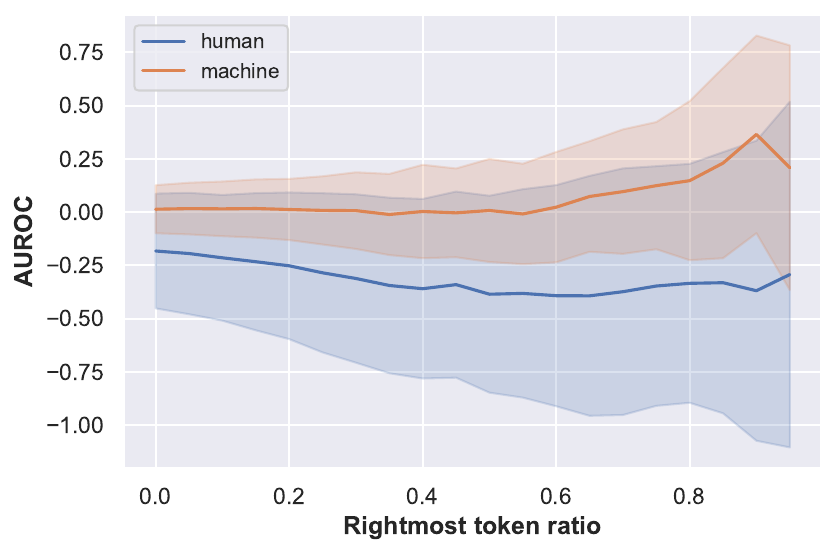}
\caption{Influence of ratio $\gamma$ on the curvature mean and standard deviation plot. }
\label{fig: deviation}
\end{figure}

\begin{figure}[ht]
\centering
\includegraphics[width=1.\linewidth]{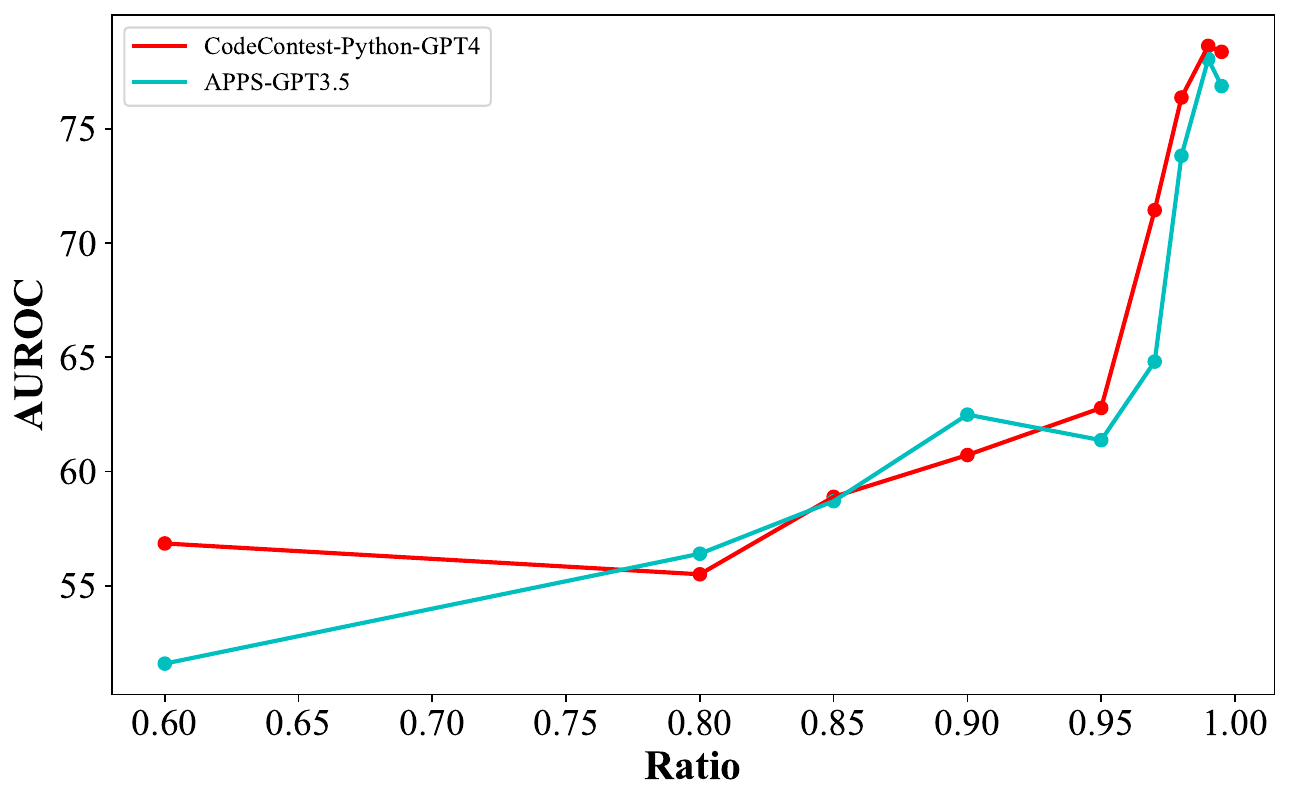}
\caption{Influence of ratio $\gamma$ on the detection AUROC score of DetectGPT4code. }
\label{fig: ratio}
\end{figure}

Since the preceding parts of codes usually occupy the lower probability curve part of a model when the model just starts decoding and the decoding becomes more deterministic when the decoding length is moderately long, we assume using the logits of the rightmost tokens might improve the results. To validate such a hypothesis, we perform a visualization experiment by changing $\gamma$ from $0$ to $1.0$ with an interval of $5\%$ and plot the curvature mean and standard deviation values in Figure \ref{fig: deviation}. As we can see, the area of two curves overlapped region starts decreasing as $\gamma$ increases. This validates our assumption that as decoding proceeds for codes, the decoding process becomes more deterministic. The resulting classification also becomes easier since given long enough preceding codes, the ending codes are also more likely determined. Subsequently, we draw the detection results in Figure \ref{fig: ratio}.
The truncation ratio is originally picked from a wide range but we find the ignorable difference before $\gamma$ reaches 0.6, so we only visualize $\gamma \in \{0.6, 0.8, 0.85, 0.9, 0.95, 0.98, 0.99, 0.995\}$ in the plots. As we can see, as $\gamma$ increases, the detection results consistently increases, but starts decreasing after $\gamma=0.99$. This also corresponds to the visualization in Figure \ref{fig: deviation}.

\subsection{Influence of Perturbation Numbers}
\begin{figure}[ht]
\centering
\includegraphics[width=.95\linewidth]{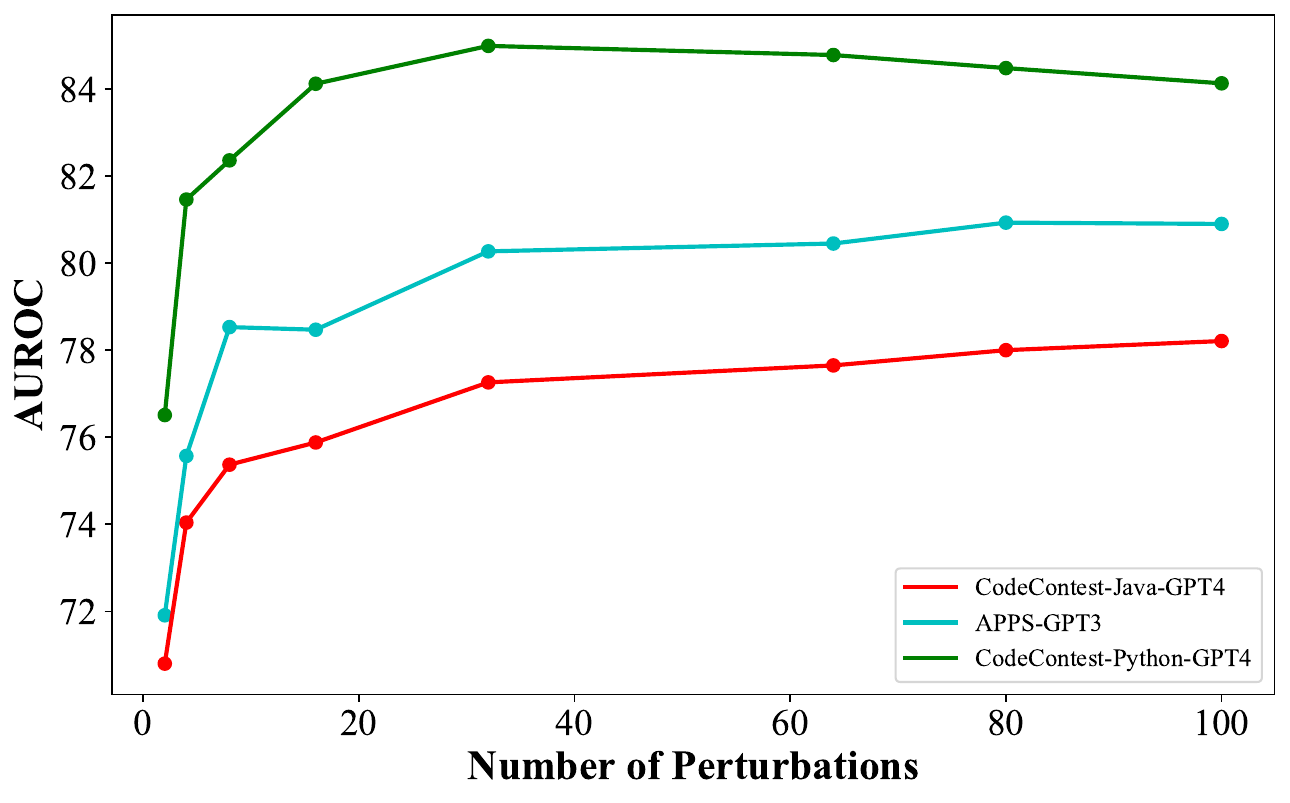}
\caption{Influence of the number of perturbations on the detection AUROC score of three scenarios. }
\label{fig: pertub}
\end{figure}

To test how the number of perturbations influences the detection performance, we plot the AUROC score in Figure \ref{fig: pertub} by varying the perturbation numbers in $\{2, 4, 8, 16, 32, 64, 80, 100\}$.
In general, we can see an increasing trend since more perturbations lead to better discriminative ability, though requiring more computational cost. As we can see, the detection quality continues increasing until 100 perturbations, while the marginal gain decreases. Typically, when the number of perturbations is under 32, the detection quality drops significantly. Intuitively, under the assumption that machine-generated text usually lies in the negative curvature regions, more perturbations can erase the possibilities that human-written text also lies in the negative regions, thus making the discrimination easier.
So we believe end users can balance the quality and effectiveness by choosing an appropriate number of perturbations.

\subsection{Influence of FIM Model Size}
\begin{figure}[ht]
\centering
\includegraphics[width=1.\linewidth]{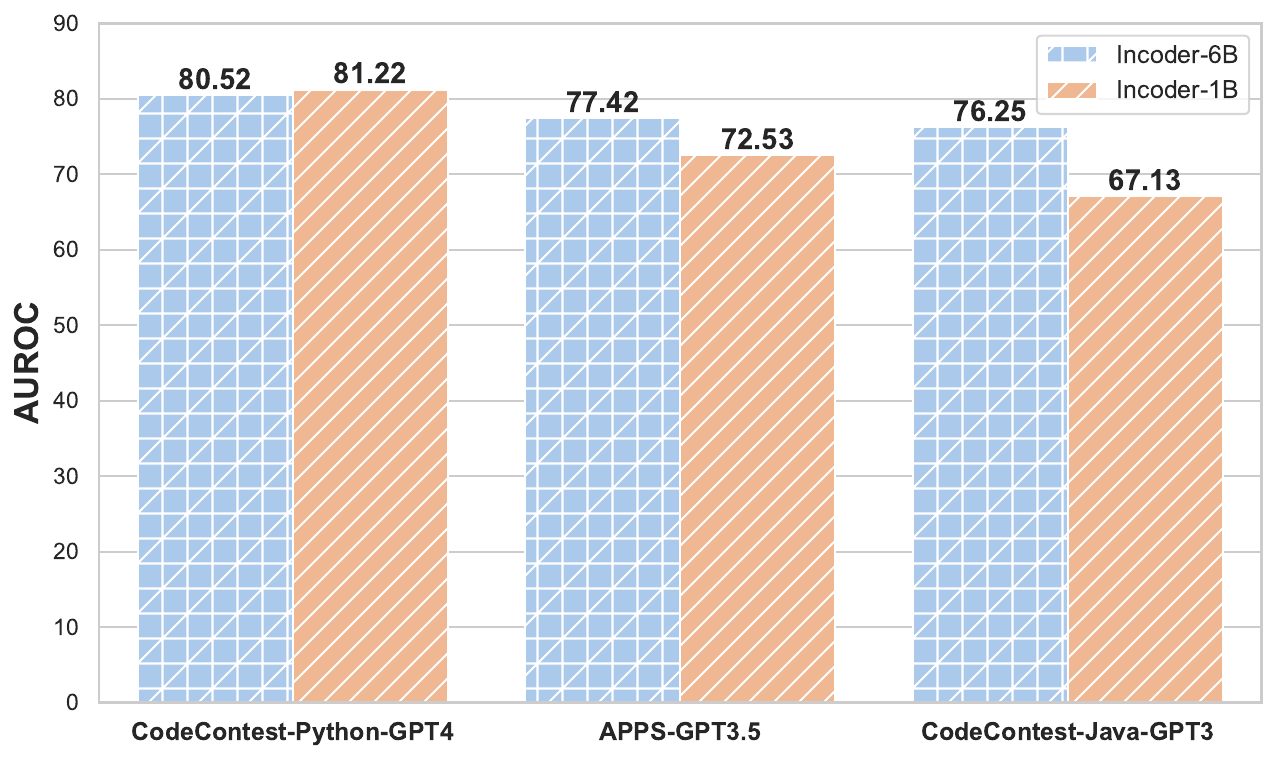}
\caption{Influence of fill-in-the-middle model size on the detection AUROC. }
\label{fig: size}
\end{figure}

Here, we show how detection performance varies as the surrogate model changes from Incoder-6B to Incoder-1B in Figure \ref{fig: size}. It is clear that on both APPS and JAVA, the 6B model always achieves significantly better results than the smaller 1B counterpart, though there is little difference in the CodeContest-Python result.
We suspect a larger model can perform better FIM tasks, resulting in improved perturbations. 

\subsection{Influence of Surrogate model}
\begin{figure*}
\centering
    \includegraphics[width=.9\textwidth]{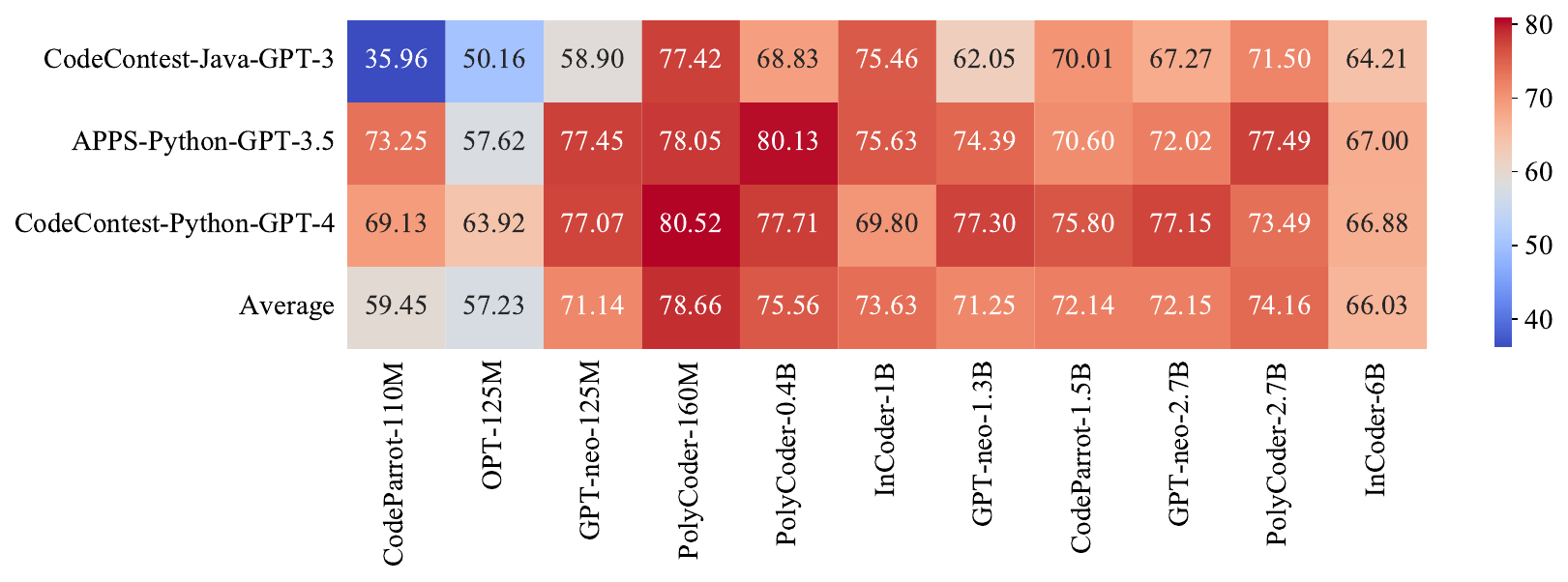}
    \caption{A heatmap comparison of detection AUROC scores with the increasing surrogate model sizes, ranging from 110M on the leftmost to 6B on the rightmost. }\label{fig: surrogate}
\end{figure*}

Here, we show how the different surrogate models affect the detection quality. In our experimentation, we employed a diverse range of surrogate models, encompassing model families of various sizes. These included several million-level models such as OPT-125M, CodeParrot-110M, PolyCoder-160M, and PyCodeGPT, as well as several billion-level models such as GPT-neo-2.7B and Incoder-6B. The results are summarized in Figure \ref{fig: surrogate}. As we can see from this heatmap, OPT-125M performs the worst, possibly due to the lack of training data on codes. This is in contrast to the conclusion \citep{mireshghallah2023smaller} that OPT-125M is the best universal detector in text detection. On the other hand, code language models like the smallest code model CodeParrot-110M achieve the worst result on Java, which is expected since it is only trained on python codes. But larger models like PolyCoder-2.7B, Incoder-6B also achieve suboptimal results, while PolyCoder-160M achieves universal best results on three datasets. Therefore, we make the conclusion that a smaller code language model is a better universal code detector if it is specifically trained on the target programming languages.

\subsection{Effect of FIM Lines}
This section investigates the impact of the number of fill-in-the-middle (FIM) masked lines on detection quality. For the code infilling task, we employ Incoder-6B to fill 2, 4, 8, and 12 masked lines. It should be noted that in cases where instances consist of less than 10 lines, we mask their corresponding maximum lines. Furthermore, we limit the number of masked lines to 12 for consistency.
Table \ref{tab: line} presents a summary of the results. Overall, masking 8 lines yields the optimal strategy as both fewer and more masked lines negatively affect the AUROC and TPR. Intuitively, masking only a few lines fails to introduce perturbations that deviate significantly from the original code, while masking too many lines deviates excessively from the original code. We would like to emphasize that this conclusion is based on our analysis of the tested code corpora and the optimal number of masked lines may vary for other datasets with a significantly different number of lines of code.

\begin{table*}[ht]
\centering
\resizebox{0.88\textwidth}{!}
{
\begin{tabular}{lcccccccc}
\toprule
\# FIM Lines & \multicolumn{2}{c}{2} & \multicolumn{2}{c}{4} & \multicolumn{2}{c}{8} & \multicolumn{2}{c}{12} \\
\cmidrule(lr){2-3} \cmidrule(lr){4-5} \cmidrule(lr){6-7} \cmidrule(lr){8-9}  
Dataset\&Model & AUROC & TPR & AUROC & TPR & AUROC & TPR & AUROC & TPR  \\
\midrule
CodeContest-Java-GPT3.5 & 58.57 & 18.58 & 62.11 & 22.12 & \textbf{64.03} & \textbf{28.32} & 62.79 & 15.04 \\
\hline
APPS-GPT3 & 59.40 & 21.21 & 73.30 & 32.57 & \textbf{79.89} & 44.63 & 79.02 & \textbf{47.82} \\
APPS-GPT4 & 64.47 & 17.78 & 66.98 & 34.82 & 76.25 & \textbf{52.29} & \textbf{77.90} & 45.93 \\
\hline
CodeContest-Python-GPT3 & 47.44 & 9.52 & 73.30 & 32.57 & \textbf{79.89} & \textbf{44.63} & 73.80 & 20.95
\\
\bottomrule
\end{tabular}

}
\caption{As we increase the number of FIM lines, both the detection AUROC and TPR increase. Nevertheless, we observed a degradation in performance when extending beyond 8.
}\label{tab: line} 
\end{table*}

\section{Challenges}
\subsection{Programming Languages}
According to the findings presented in Table \ref{tab:overall} and Table \ref{tab: line}, switching to Java as the programming language consistently leads to a decline in performance. Additionally, the heatmap illustrated in Figure \ref{fig: surrogate} provides further evidence of the degradation observed across nearly all surrogate models. Notably, CodeParrot, which is not explicitly trained in Java, yields exceptionally low results. Although other surrogate models are trained on Java code, they still achieve suboptimal outcomes. Given the diverse range of programming languages, developing a universal code detector capable of handling all languages poses a significant challenge. To address this, we recognize the need for more extensive experimentation in future studies.

\subsection{Revised Codes Attack}
To count for the real scenario where humans tend to modify machine-generated codes, we conduct experiments to investigate the detection of revised codes attack.
We use the powerful Incoder-6B model to perform revision by randomly masking 2 or 4 lines and then refilling them. Considering that masking additional lines would significantly deteriorate the quality of the generated codes, it is less likely to be a practical approach for adoption. 
The results are reported in Table \ref{tab: attack}. As we can see, although both detection performance decreases, the overall drop is not significant. More sophisticated revisions are left for future work.

\begin{table}[ht]
\centering
\resizebox{0.48\textwidth}{!}
{
\begin{tabular}{lcccc}
\toprule
Misc & \multicolumn{2}{c}{Revised} & \multicolumn{2}{c}{Drop $\downarrow$ }  \\
\cmidrule(lr){2-3} \cmidrule(lr){4-5} 
Dataset & AUROC & TPR & AUROC & TPR \\
\midrule
APPS-GPT4, revised 4 lines & 73.16 & 47.78 & 3.09 & 4.51 \\
\hline
APPS-GPT3, revised 2 lines  & 77.40 & 41.21 & 2.49 & 3.42  \\
\bottomrule
\end{tabular}

}
\caption{Revised attack: our detector witnesses slight performance drop when 2 or 4 lines in the original codes are modified.
}\label{tab: attack} 
\end{table}\vspace{-0.3cm}

\section{Conclusion}

We find that both previous training-based and zero-shot text detectors could not achieve satisfactory results on codes. Then, we propose DetectGPT4Code for code detection and achieve new SOTA results on various benchmarks. 
With the rapid advance of LLMs for code generation, we hope our research provides timing solutions to this new challenge. 
In the future, this task will become even harder as the LLMs continue revolutionizing their code generation ability. Therefore, we call for more focus on this new direction from the whole community.

\section*{Limitations}
Although we try our best to explore the detection of machine-generated codes, we notice the main limitation of our work falls into two folds: first, our work only explores several LLMs, but there are many other LLMs that we have not tested on, and we believe extensively testing on all of them falls out of the scope of this work, so we leave them for future work.
Second, the overall code detection results are still lagging behind text detection by a substantial margin. But we argue that this topic is relatively new and underexplored, so as the preliminary exploration, we believe our methodologies are strong enough to serve as competitive baselines for future research, especially considering the increased difficulty of code detection.

\section*{Ethics Statement}
We want to claim that the detection results do not represent the personal opinions of the authors, and potential users, especially teachers, should take their own risks relying on the detection conclusion. We want to point out that the current detection could still make mistakes, and possibly show bias towards certain models or programming languages. With those said, we hope our research provides some insights into this challenging new task, but users should be responsible for themselves.


\bibliography{anthology,custom}
\bibliographystyle{acl_natbib}

\appendix

\section{Appendix}
\subsection{Baselines}\label{app: baseline}
In DNA-GPT, we follow their recommended settings with 10 times re-prompting with $\gamma = 0.5$ for all experiments. Although increasing the number of re-prompting might increase the performance, we stick to 10 due to budge limit.

In DetectGPT, we replace T5-large by Incoder-6B and generate 20 perturbations, in consistency with our own setting.

\subsection{Code extraction}
Following \citep{chen2021evaluating}, we terminate the sampling process if one of the following sequences
is encountered in the generation process: `\verb|\nclass|', `\verb|\ndef|', `\verb|\n#|', `\verb|\n@|', `\verb|\nif|', and `\verb|\nprint|'. In practice, we also use the GitHub codes from HumanEvalPlus \citep{liu2023your} to do such filtering.

\label{sec:appendix}

\end{document}